\documentclass[10pt,twocolumn,letterpaper]{article}

\usepackage{iccv}
\usepackage{times}
\usepackage{epsfig}
\usepackage{graphicx}
\usepackage{subcaption}
\usepackage{amsmath}
\usepackage{amssymb}

\usepackage[pagebackref=true,breaklinks=true,letterpaper=true,colorlinks,bookmarks=false]{hyperref}
\usepackage{bookmark}
\usepackage{booktabs}
\usepackage{diagbox}
\usepackage[export]{adjustbox}

\iccvfinalcopy 

\def\iccvPaperID{****} 

\ificcvfinal\fi
\begin{document}

\title{Style2Vec: Representation Learning for Fashion Items from Style Sets}

\author{
    Hanbit Lee, Jinseok Seol, Sang-goo Lee \\
    Department of Computer Science and Engineering \\
    Seoul National University \\
    {\tt\small \{skcheon, jamie, sglee\}@europa.snu.ac.kr}
}

\maketitle

\begin{abstract}
   With the rapid growth of online fashion market, demand for effective fashion recommendation systems has never been greater. In fashion recommendation, the ability to find items that goes well with a few other items based on style is more important than picking a single item based on the user's entire purchase history. Since the same user may have purchased dress suits in one month and casual denims in another, it is impossible to learn the latent style features of those items using only the user ratings. If we were able to represent the style features of fashion items in a reasonable way, we will be able to recommend new items that conform to some small subset of pre-purchased items that make up a coherent style set.
We propose \textit{Style2Vec}, a vector representation model for fashion items. Based on the intuition of distributional semantics used in word embeddings, Style2Vec learns the representation of a fashion item using other items in matching outfits as context. Two different convolutional neural networks are trained to maximize the probability of item co-occurrences. For evaluation, a fashion analogy test is conducted to show that the resulting representation connotes diverse fashion related semantics like shapes, colors, patterns and even latent styles. We also perform style classification using Style2Vec features and show that our method outperforms other baselines.
\end{abstract}

\section{Introduction}
\begin{figure}[!htp]
    \centering
    \begin{subfigure}[b]{0.32\linewidth}
        \centering
        \includegraphics[width=0.95\textwidth,valign=b]{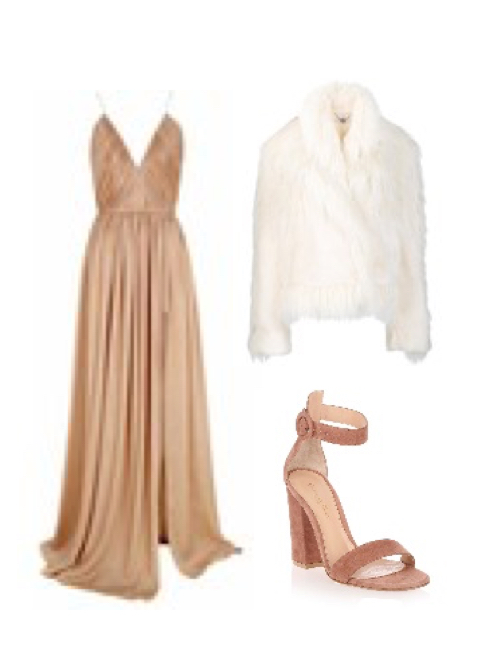}
        \caption{\label{sfig:example-formal}}
    \end{subfigure}
    \begin{subfigure}[b]{0.32\linewidth}
        \centering
  \includegraphics[width=0.95\textwidth,valign=b]{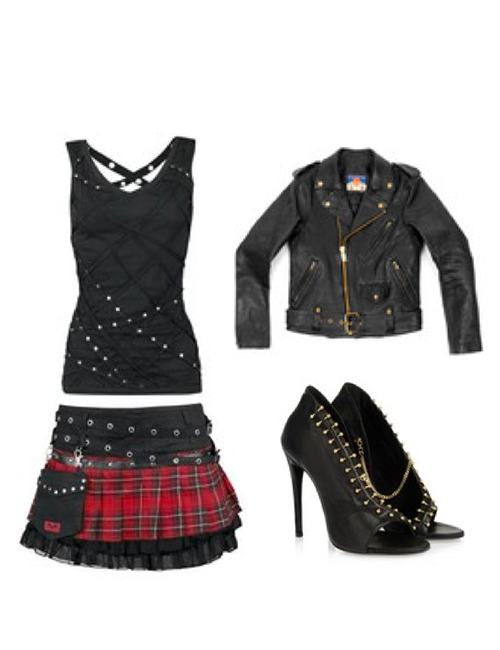}
        \caption{\label{sfig:example-punk}}
    \end{subfigure}
    \begin{subfigure}[b]{0.32\linewidth}
        \centering
        \includegraphics[width=0.95\textwidth,valign=b]{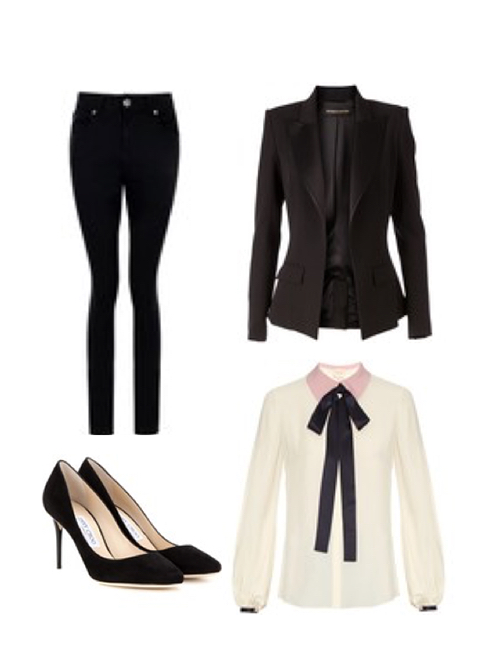}
        \caption{\label{sfig:example-business}}
    \end{subfigure}
    \caption{Examples of style sets in (a) formal, (b) punk and (c) business types. A style set is collage of coherent items from distinct categories.\label{fig:example}}
\end{figure}
In fashion recommendation, the ability to find items that \textit{goes well} with a few of other items based on style is more important than recommending a single item without context. Traditional recommendation approaches use user purchase data (generally referred to as \textit{user ratings}) to learn the latent features of item sets. However, since the same user may have purchased both dress suits and casual denims (within a certain time span), it is impossible to learn specific features of individual styles using such user ratings. Therefore, the ability to identify detailed style features of an item is quite challenging. A straightforward approach is to use item categories and hand-crafted attributes to represent an item. However, it is almost impossible to define a set of fine-grained attributes exquisite enough to characterize the subtle nuisance of fashion. Recent approaches use visual images of items to obtain more rich latent item features. However, individual fashion features cannot effectively explain whether one item matches another or a set of items make up a stylish outfit.

In this paper, we propose Style2Vec, a distributed representation of fashion items that is learned from a large dataset of user created ``style sets''. We define a \textit{style set} to be a collection of garments and accessories that make up a single outfit. Figure \ref{fig:example} shows three examples of style sets in different types. As we can see, a style set is composed of items belong to different categories and our network attempts to learn latent style shared by items in distinct shapes.

To devise a model that can learn styles from set of item images, we borrow the intuition of distributional semantics from the natural language processing community. Word2Vec \cite{mikolov2013distributed} and other distributed word representation models learn the representation by training shallow neural network to maximize the probability of word co-occurrences within a context window (or sentence). For Style2Vec, a word is a single fashion item, (e.g., a jacket, a shirt, a hat, etc.) while a sentence corresponds to a style set. The key idea of our approach is to use the member items of a style set as the context items. While Word2Vec model directly updates word embedding vectors, but for fashion items, it is impossible to learn embedding vectors directly since each item appears in small number of style sets. To overcome such context sparseness, we use convolutional neural network (CNN) a as projection network from image to embedding vector since CNN clumps similar images. This utilizes even rare items as meaningful vectors, which was difficult in traditional Word2Vec models.

Total 297,083 user created style sets have been collected from a popular fashion website as our corpus for training and testing. We use two different convolutional neural networks to map an image to its latent features, one for the input item and the other for the context items. The former network, after training, is used as the embedding network for extracting the latent style feature vector for each item.

For evaluation, a fashion analogy test is conducted to show that the resulting representation connotes diverse fashion related semantics like shapes, colors, patterns and even latent styles. We also perform style classification using Style2Vec and show that our method outperforms other baselines.

\section{Related Work}
There are several works trying to uncover visual relationship between pair of compatible fashion items. McAuley et al.~\cite{mcauley2015image} use parameterized distance metric (i.e. Mahalanobis distance) to learn relationships between co-purchased item pairs. He et al. \cite{he2016learning} extends \cite{mcauley2015image} to allow multiple notions of `relatedness' with ensemble of $k$ Mahalanobis transforms. Both of these approaches only use image features extracted from the network that is trained for image classification of general images. So, the item features they use are not fashion-centric nor context-sensitive.

The closest work to ours is the work by Veit et al. \cite{veit2015learning}. They use Siamese convolutional neural network to learn dyadic item co-occurrences. Since they train network to minimize the $l_2$ difference between co-purchased pair of item features, it is impossible for them to learn features from set of arbitrary number of items. In addition, they use co-purchase dataset from Amazon\footnote{https://www.amazon.com/} to learn latent features of fashion items. However, as mentioned before, we claim that co-purchase data is not suitable for identifying specific style relationships among well-matched style set.

\begin{figure}[htp]
    \centering
    \adjustbox{minipage=1.5em,valign=c}{\subcaption{}\label{sfig:model-vgg}}%
    \begin{subfigure}[c]{\dimexpr1.0\linewidth-3em\relax}
        \centering
        \includegraphics[width=0.95\textwidth]{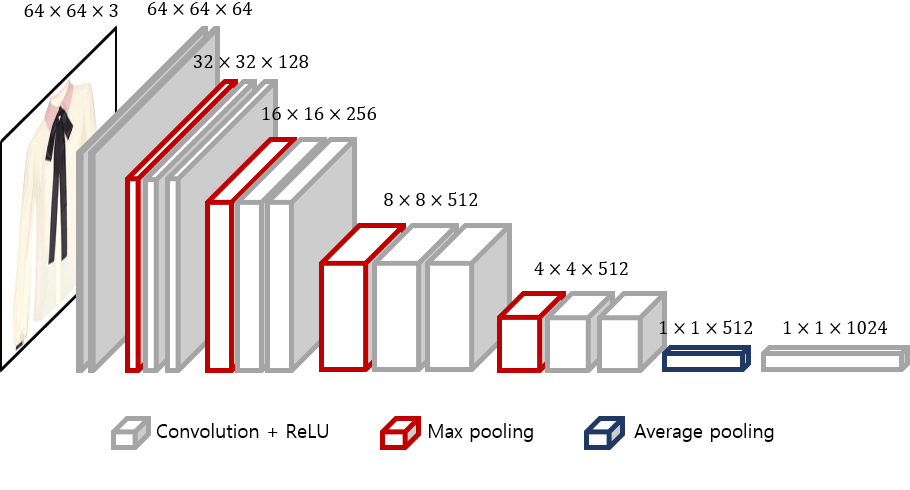}
    \end{subfigure}
    \adjustbox{minipage=1.5em,valign=c}{\subcaption{}\label{sfig:model-skipfashion}}%
    \begin{subfigure}[c]{\dimexpr1.0\linewidth-3em\relax}
        \centering
        \includegraphics[width=0.95\textwidth]{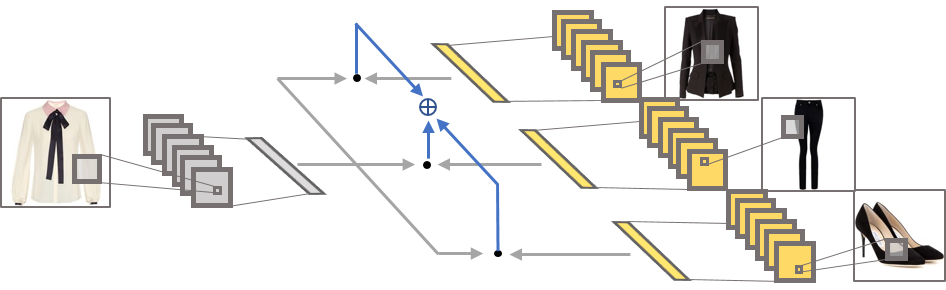}
    \end{subfigure}
    \caption{Model architecture of Style2Vec. (a) Structure of VGGNet \cite{simonyan2014very} that is used as input and context projection networks. (b) Process of computing the loss function given a style set. CNN colored in gray is VGGNet for the input item, and the yellow one is for the context items.\label{fig:model}}
\end{figure}

Another research stream we need to consider is unsupervised representation learning from images. In this area, generative adversarial networks \cite{goodfellow2014generative} are receiving enormous attention as they can generate highly realistic images from latent vectors and they have been shown to learn good feature representation from images \cite{donahue2016adversarial, dumoulin2016adversarially, radford2015unsupervised}. The most popular architecture is DCGAN \cite{radford2015unsupervised}, which uses transposed convolutional network for generator and convolutional network for discriminator that are trained in adversarial fashion. However, these approaches learn representation from standalone image, not in the context of relationship among items in a set.

Meanwhile, recently, noteworthy fashion recommender systems are being proposed, which try to recommend items that go well with other items. Lee et al. \cite{lee2015style} use heterogeneous graph to link fashion items those make up a stylish outfit, as well as to link items to their attributes. Hu et al. \cite{hu2015collaborative} propose a tensor factorization approach to recommend a set of fashion items. They do not learn item features based on sets, but use discrete item attributes or low level image features.

In summary, our approach is distinct from the above in that we aim to learn latent styles of fashion items from style sets consist of multiple items. Our approach captures fashion semantics on the style space we learn, which can be effectively utilized by fashion recommender systems.

\section{Style2Vec model}
Our goal is to implement a representation learning framework for fashion items where learned representation contains latent styles shared by items in a style set. The latent styles of an item are highly determined by the other items in the same style set as well as by features of item itself. In order to utilize features of the context items, we borrow the concept of distributional hypothesis from linguistics, meaning that the words used and occur in the similar contexts tend to purport similar meanings. We assume that fashion items in the similar style set share a coherent style, as words in the similar contexts have similar meanings.

Many researches in natural language processing have tried to effectively represent semantic of a word using the words in the same context under this hypothesis. Most successful model is Word2Vec model \cite{mikolov2013linguistic} which learns a simple neural network to reconstruct linguistic contexts of words. The network has one hidden layer and there are two types of parameters, one for projecting input word to its latent vector on the hidden layer and another for multiplying weight vector to the latent vector so as to predict distribution of context words.

Inspired by Word2Vec model, we devise our model \emph{Style2Vec} to use two types of neural network, one for projecting a target fashion item to its representation vector and another for mapping the other context items in the same style set to their weight vectors. The inner product between target item vector and each context weight vector is made to probabilities with softmax function and entire networks are trained to maximize the sum of the log probabilities. Figure~\ref{fig:model} shows our entire model architecture.

Formally, let the image of an item $i$ be $I_i$. We parameterize two types of networks $f_{\text{in}} : I_i \mapsto u_i$ for the input item image $I_i$ and $f_{\text{cxt}} : I_c \mapsto v_c$ for the context item image $I_c$, both from the image space $\mathcal{I}$ to a latent feature space $\mathbb{R}^d$. We use the same VGGNet \cite{simonyan2014very} structure for both networks, except that they do not share parameters since those two networks have different purposes. Our goal is to learn $f_{\text{in}}$ and $f_{\text{cxt}}$ that maximize the log probability of the item co-occurrence in a style set.

Let $F$ be a set of all fashion items. We assume that each of \emph{style set} $S$ in our training dataset $D \subset \mathcal{P}(F)$ has coherent style. We iterate $i$ over $S$ as an input, with items in $S \setminus \{i\}$ as a context. The overall objective of Style2Vec model is to maximize the average log probability for each $S$ in $D$:

\begin{equation}
\frac{1}{|S|} \sum_{i \in S} \sum_{c \in S \setminus \{i\}} \log \left( \frac{\exp(u_i^\intercal v_c)}{\sum_{j \in F} \exp(u_i^\intercal v_j)} \right)
\end{equation}

\noindent
where $u_i = f_{\text{in}}(I_i)$, $v_c = f_{\text{ctx}}(I_c)$ and $v_j = f_{\text{ctx}}(I_j)$.

Since computing the softmax is very expensive, we apply negative sampling approach \cite{mikolov2013distributed}. We randomly pick $k$ items, different from the items in the training sample $S$ to form a negative sample set $N_{S, i, c} \subset F \setminus S$ for each $(i, c)$. Now our objective becomes

\begin{equation}
\frac{1}{|S|} \sum_{i \in S} \sum_{c \in S \setminus \{i\}} \left( \log \sigma (u_i^\intercal v_c) + \sum_{j \in N_{S, i, c}} \log \sigma (-u_i^\intercal v_j) \right)
\end{equation}

\noindent
for each $S$ in $D$.

\begin{figure*}[!htbp]
  \includegraphics[width=\textwidth]{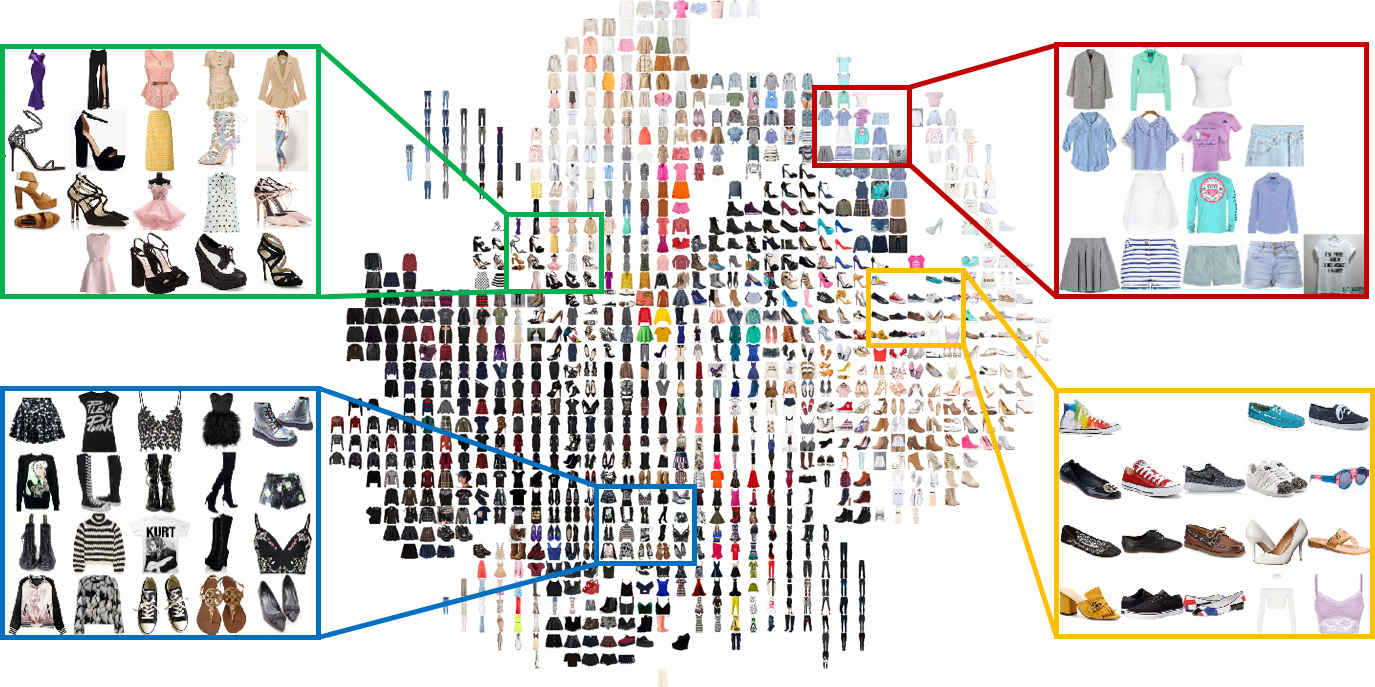}
  \caption{Visualization of style space in 2D using t-SNE \cite{maaten2008visualizing}. Upper left box shows that the items with formal style types are clustered. Similarly, bottom left box can be seen as a cluster of punk styled items. Meanwhile, items in upper right box share subtle pastel color tones and the casual shoes are grouped together in bottom right box.\label{fig:tsne}}
\end{figure*}

It is noteworthy that Style2Vec is capable of learning any size of a style set, since we use shared parameters for projecting all the context item images to their context features.


\section{Experiment}
To evaluate the item representation of Style2Vec, we conducted several tasks using the item vectors learned by Style2Vec. Visualization of Style2Vec feature space shows how items with similar style gather together. Also we investigate the characteristics of item representation by performing a fashion analogy test. Experiment of style type classification demonstrates that our item vectors can be effectively used for more general fashion related tasks.

\subsection{Dataset and training}

We collected 297,083 sets from popular fashion web service Polyvore\footnote{http://www.polyvore.com/}. Each of set is composed with 2 to 4 categorically distinct items from a pool of 53,460 tops, 43,180 bottoms, 31,199 outers, 77,981 pairs of shoes, and 30,852 dresses, 236,672 unique items in total.
We use VGGNet with 16 layers including 4 max pooling layers and one average pooling layer, which maps an image to 1,024 dimensional vector. The whole network is trained using mini-batch gradient descent through Adam optimizer \cite{kingma2014adam} with batch normalization \cite{ioffe2015batch}.

\subsection{Visualization of style space}

The t-SNE \cite{maaten2008visualizing} algorithm is used to embed 1,024 dimensional item features into a 2D space. Figure~\ref{fig:tsne} shows visualized style space of our model. Since the convolutional neural network tend to map visually similar items to close points in the latent space, we can observe that the items of similar shape and color (e.g., black mini-skirt) are projected nearby. However, even when images are not visually similar, they might lie near together if they have been matched with similar context items. We can observe that the pastel tone items those go well with each other are located close to each other (upper right box). Also, sneakers and casual flat shoes gather together, since those items make good matches with other casual items (lower right box). It is remarkable that formal dress and heels are clustered, as well as the punky items of distinct categories are located nearby (upper left box \& lower left box). This visual map gives us a glimpse of our model and justifies that our model transforms the images into a reasonable style space.

\begin{figure}[!htbp]
  \adjustbox{minipage=1.5em,valign=c}{\subcaption{}\label{sfig:analogy-a}}%
  \begin{subfigure}[c]{\dimexpr1\linewidth-2em\relax}
    \centering
    \includegraphics[width=\textwidth,valign=c]{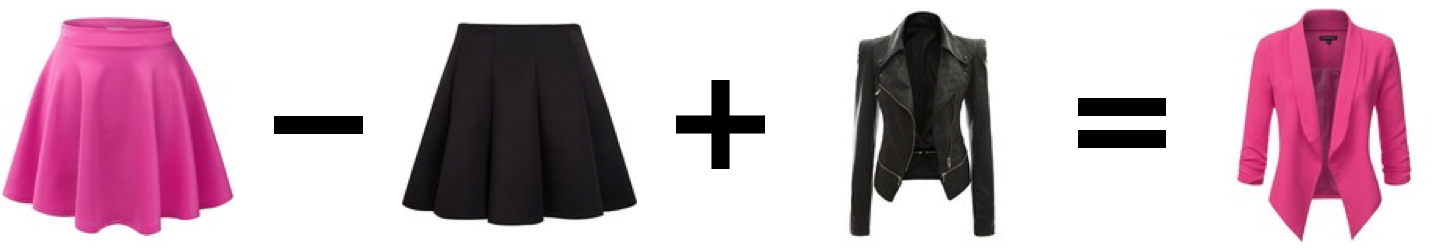}
  \end{subfigure}
  \adjustbox{minipage=1.5em,valign=c}{\subcaption{}\label{sfig:analogy-b}}%
  \begin{subfigure}[c]{\dimexpr1\linewidth-2em\relax}
    \centering
    \includegraphics[width=\textwidth,valign=c]{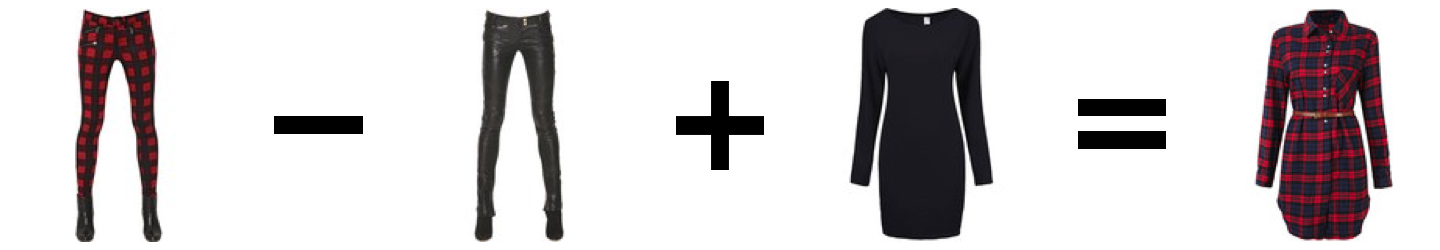}
  \end{subfigure}
  \adjustbox{minipage=1.5em,valign=c}{\subcaption{}\label{sfig:analogy-c}}%
  \begin{subfigure}[c]{\dimexpr1\linewidth-2em\relax}
    \centering
    \includegraphics[width=\textwidth,valign=c]{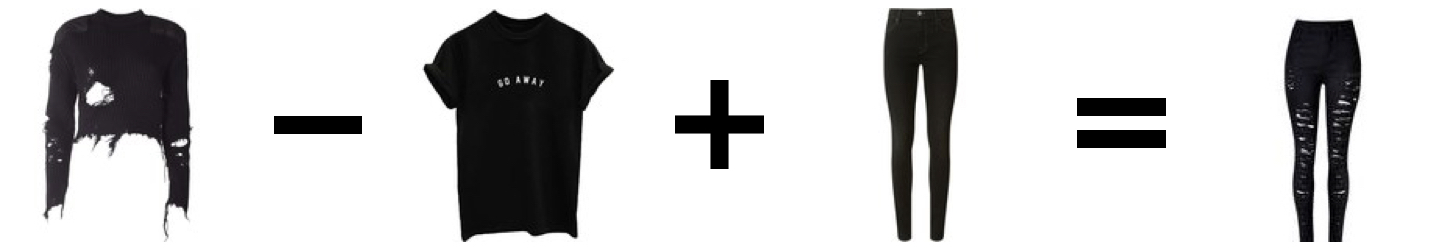}
  \end{subfigure}
  \adjustbox{minipage=1.5em,valign=c}{\subcaption{}\label{sfig:analogy-d}}%
  \begin{subfigure}[c]{\dimexpr1\linewidth-2em\relax}
    \centering
    \includegraphics[width=\textwidth,valign=c]{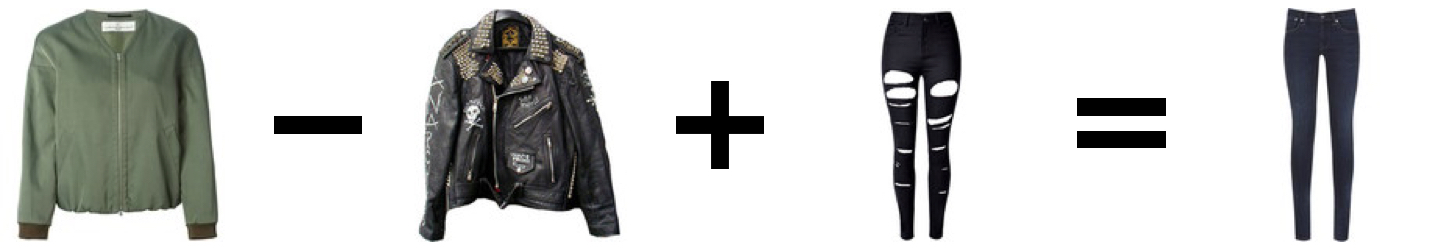}
  \end{subfigure}
  \adjustbox{minipage=1.5em,valign=c}{\subcaption{}\label{sfig:analogy-e}}%
  \begin{subfigure}[c]{\dimexpr1\linewidth-2em\relax}
    \centering
    \includegraphics[width=\textwidth,valign=c]{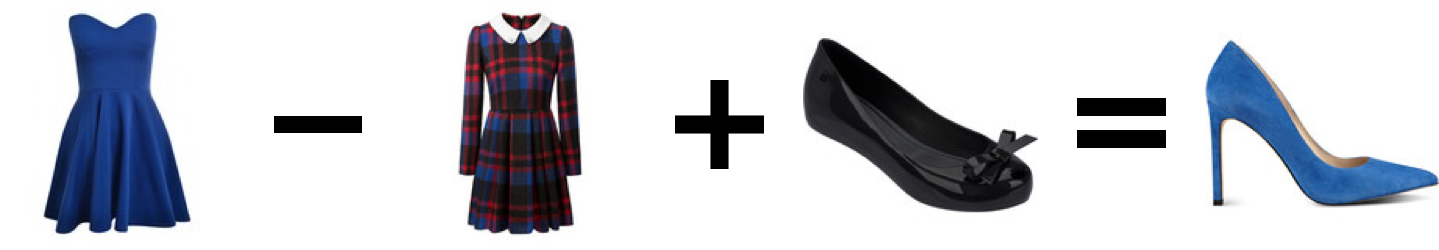}
  \end{subfigure}
  \adjustbox{minipage=1.5em,valign=c}{\subcaption{}\label{sfig:analogy-f}}%
  \begin{subfigure}[c]{\dimexpr1\linewidth-2em\relax}
    \centering
    \includegraphics[width=\textwidth,valign=c]{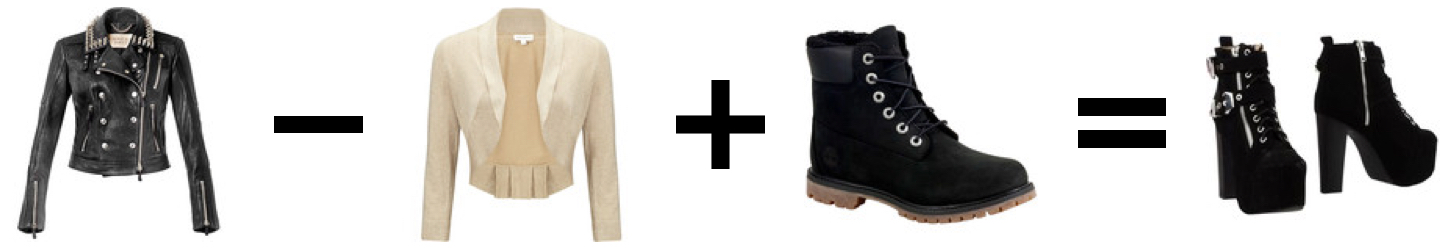}
  \end{subfigure}
  \caption{Results from analogy tests. Style factors can be calculated using basic arithmetics. Each test checks the transfer of following fashion factor: (a) color (b) pattern (c) adding punk style (d) removing punk style (e) adding formal style (f) adding punk style.\label{fig:analogy}}
\end{figure}

\subsection{Analogy test}

In order to investigate the inherent characteristics of latent style features, we performed a fashion analogy test.
Each test question has a form of ``$x$ is to $y$ as $z$ is to $\rule{0.2cm}{0.15mm}$'' which asks the system to find an item that fills the blank, in the context of fashion semantics.
For example, a question ``punk boots is to punk jacket, as formal heels is to $\rule{0.2cm}{0.15mm}$'' (answer might be a formal blazer) can test whether there exist some regularities related to latent styles. We manually created 288 test questions. Specifically, we selected two items $x$ and $y$ from a style set, where $x$ and $y$ belong to different categories but sharing a concrete style (e.g., punk boots and punk jacket). For $z$, we choose a random item that belongs to same category as $x$ (e.g., formal heels).
To fill the blank, we first obtain the item feature vectors of $x$, $y$, and $z$ with our trained model. Then we find the answer item by calculating the closest item vector to the vector $u_y - u_x + u_z$, where $u_x$, $u_y$, $u_z$ are feature vectors of item $x$, $y$ and $z$ \cite{mikolov2013efficient}.
The answer item found by the system is evaluated by two human evaluators only if the answer item belongs to the same category as $y$. If the category of the answer item is different from that of $y$, then we mark that question as failed one. Furthermore, we mark as \emph{acceptable} only if both of two human evaluators agree together that $y$ and the answer item go well enough to form a style set. Total 199 out of 288 questions are marked as acceptable, which is about 69.1\%. This result quantitatively supports that our style features possess latent styles shared within a style set.

Among the analogy tests, we could find remarkable results which show that the feature vectors of our model possess basic characteristics such as color, pattern, shape as well as latent styles. Moreover, those latent factors can be transfered via simple vector arithmetic.
Figure~\ref{fig:analogy} shows some of the results where the first three items of each row correspond to items $y$, $x$, $z$ in order, and the rest item in the same row corresponds to the answer item found by the system.
Cases of (a) and (b) presents test results related to color and pattern factor.
In case of (a), $x$ and $y$ are items with same category and detailed shape, but only different in color (briefly, a black mini skirt and a pink mini skirt), while $z$ is an item in a different category but in same color as $x$ (which is a black blazer).
The model finds the pink blazer, which implies that the color factor is transferred via simple vector arithmetic.
Similar regularity was found in case of pattern factor from (b).
Cases of (c) and (d) show style transfer between punk style and casual style.
In case of (e), we can observe transfer of formal style from formal dress to high-heels via casual dress and flat shoes.
Moreover, we could find blue high heels those match well with given blue formal dress, as if it were a recommendation.

The result of analogy tests implies that our embedding model can learn fashion features of color, shape, pattern, and even latent styles, which are considered as the most ambiguous property in the fashion industry. We claim that our latent style features are derived from process of learning items in a style set simultaneously.

\begin{table}
  \begin{tabular}{c|cccc|c}
    \toprule
    Set size & Casual & Punk & Formal & Business & Total \\
    \midrule
    2 & 978 & 943 & 773 & 1,000 & 3,694 \\
    3 & 1,000 & 1,000 & 1,000 & 961 & 3,961 \\
    4 & 759 & 886 & 1,000 & 311 & 2,956 \\
    \midrule
    total & 2,737 & 2,829 & 2,773 & 2,272 & 10,611 \\
    \bottomrule
  \end{tabular}
  \caption{Distribution of style set data used for style classification. For fare comparison, we collected sets of four style types uniformly.\label{tab:data}}
\end{table}

\subsection{Style classification for style sets}

We try to classify style sets into explicit style types using item representation learned from large dataset of style sets. Style features of a style set is obtained by averaging the style vectors of items in the set. The style vector of each fashion item is achieved from three different models: Siamese CNN \cite{veit2015learning}, DCGAN \cite{radford2015unsupervised}, and Style2Vec.

All the models are trained on the same dataset with controlled number of parameters. However, in case of Siamese CNN, we transformed dataset by getting pairwise combinations from all the style sets, since Siamese CNN can be trained only with pairs of items. To generate negative samples, we randomly selected items from different categories to make item pairs. We also made another version of Style2Vec model by training the network on this pairwise dataset to investigate the difference between learning pairwise relationship and relationship among sets with more than two items. For DCGAN, the features are extracted from penultimate layer of discriminator CNN. So eventually item representation from four models are evaluated: Siamese CNN, DCGAN, Styl2Vec(Pair) and Style2Vec.

Since the original dataset has no style type labels on the style sets, we made a dataset for style classification task. We first choose four distinguishable style types: casual, punk, formal, and business. Then we collected style sets with specific style type by querying each style keyword to Polyvore site. Data distribution of style sets for each style type are shown in Table~\ref{tab:data}.

Overall process is as follows: four models are trained on the original large dataset and the style features of style sets in the new dataset are obtained from the trained models. 90\% of the new dataset is used for training the multi-layer perceptron classifier which classify style sets into four style types and the rest is used as a test set.

\begin{table}
  \centering
  \begin{tabular}{cc}
    \toprule
    Model & Accuracy \\
    \midrule
    Siamese CNN & 51.14\% \\
    DCGAN & 54.33\% \\
    Style2Vec(pairwise) & 54.99\% \\
    \textbf{Style2Vec} & \textbf{61.13}\% \\
    \bottomrule
  \end{tabular}
  \caption{Style classification accuracy of Siamese CNN, DCGAN, Style2Vec(Pairwise), and Style2Vec. Style2Vec(Pairwise) is a version which is trained on pairwise dataset. Multi-layer perceptron classifier is used to classify style sets to their style types. \label{tab:style-clf}}
\end{table}

Table~\ref{tab:style-clf} shows the accuracy of each model. As we can see, Style2Vec outperforms all the other models. Style2Vec performs better than DCGAN even if DCGAN is the state-of-the-art representation learning network, because DCGAN learns item representation from standalone image without considering style factors shared within a style set. Siamese CNN achieves even lower accuracy than DCGAN although Siamese CNN learns relationship of compatible item pairs. We believe there are two reasons. One reason is that, with only pairwise relationship, it is insufficient to learn the shared style type among items in a set. Another is the objective function that Siamese CNN uses is not effective enough. It is clear when we compare the results of Siamese CNN, Style2Vec trained on pairwise dataset and Style2Vec trained on original dataset. Style2Vec with pairwise dataset performs worse than the original version, which shows that it is crucial to learn relationship within a set of more than two items simultaneously. Style2Vec achieves higher accuracy than Siamese CNN even if they both use the same pairwise dataset. It shows that the objective function and the structure of Style2Vec is more effective than those of Siamese CNN.

\section{Conclusion}
In this work, we propose a novel representation learning framework, Style2Vec that can learn latent style features from large collection of user created style sets. Visualization and fashion analogy tests reveal that these style features meaningfully possess diverse semantics of shape, color, pattern and latent style. Moreover, we apply our style features to style classification task and show our approach outperforms the-state-of-the-art baselines. We believe this is a significant step towards utilizing latent style features of fashion items in various tasks including visual search, article tagging, and recommendation. As a future work, we plan to implement recommender system that recommends fashion items those go well with items in user's closet using Style2Vec item representation. We also plan to further investigate and interpret the relationships between items in a style set.

{\small
\bibliographystyle{ieee}
\bibliography{egbib}
}

\end{document}